\documentclass[letterpaper]{article} 
\usepackage{aaai25}  
\usepackage{times}  
\usepackage{helvet}  
\usepackage{courier}  
\usepackage[hyphens]{url}  
\usepackage{graphicx} 
\usepackage{multirow,multicol,makecell,booktabs,amssymb,amsmath}
\usepackage{natbib}  
\usepackage{caption} 
\usepackage{algorithm}
\usepackage{algorithmic}

\usepackage{newfloat}
\usepackage{listings}
\DeclareCaptionStyle{ruled}{labelfont=normalfont,labelsep=colon,strut=off} 
\floatstyle{ruled}
\newfloat{listing}{tb}{lst}{}
\floatname{listing}{Listing}
\title{Adaptive Learning of Consistency and Inconsistency Information\\ for Fake News Detection}
\author{
    Aohan Li \textsuperscript{\rm 1},
    Jiaxin Chen \textsuperscript{\rm 2}\thanks{Corresponding author: Jiaxin Chen (jxchen@csust.edu.cn).},
    Xin Liao \textsuperscript{\rm 1}$^{*}$,
    Dengyong Zhang \textsuperscript{\rm 2}
}
\affiliations{
    \textsuperscript{\rm 1}The College of Computer Science and Electronic Engineering, Hunan University, Changsha 410082, China.\\

    \textsuperscript{\rm 2}The School of Computer and Communication Engineering, Changsha University of Science and Technology, Changsha 410114, China.
}

\usepackage{bibentry}

\begin{document}

\maketitle

\begin{abstract}
The rapid advancement of social media platforms has significantly reduced the cost of information dissemination, yet it has also led to a proliferation of fake news, posing a threat to societal trust and credibility. Most of fake news detection research focused on integrating text and image information to represent the consistency of multiple modes in news content, while paying less attention to inconsistent information. Besides, existing methods that leveraged inconsistent information often caused one mode overshadowing another, leading to ineffective use of inconsistent clue. To address these issues, we propose an adaptive multi-modal feature fusion network (MFF-Net). Inspired by human judgment processes for determining truth and falsity in news, MFF-Net focuses on inconsistent parts when news content is generally consistent and consistent parts when it is generally inconsistent. Specifically, MFF-Net extracts semantic and global features from images and texts respectively, and learns consistency information between modes through a multiple feature fusion module. To deal with the problem of modal information being easily masked, we design a single modal feature filtering strategy to capture inconsistent information from corresponding modes separately. Finally, similarity scores are calculated based on global features with adaptive adjustments made to achieve weighted fusion of consistent and inconsistent features. Extensive experimental results demonstrate that MFF-Net outperforms state-of-the-art methods across three public news datasets derived from real social medias.
\end{abstract}

\section{Introduction}
Recently, social media platforms like Weibo and Facebook have basically replaced the traditional information dissemination methods represented by newspapers and magazines. While these platforms provide benefits such as low-cost and real-time information transmission, they have also facilitated the rampant use of fake news. Fake news refers to deliberately fabricated information that can be proven false, and its dissemination will erode public trust in public institutions and negatively affect social security \cite{capuano2023content}. In general, identifying fake news using common sense alone is challenging. Automated fake news detection technology plays a crucial role in quickly identifying fake news and effectively mitigating their impact on public opinion.

\begin{figure}[t]
\centering
\includegraphics[width=1\columnwidth]{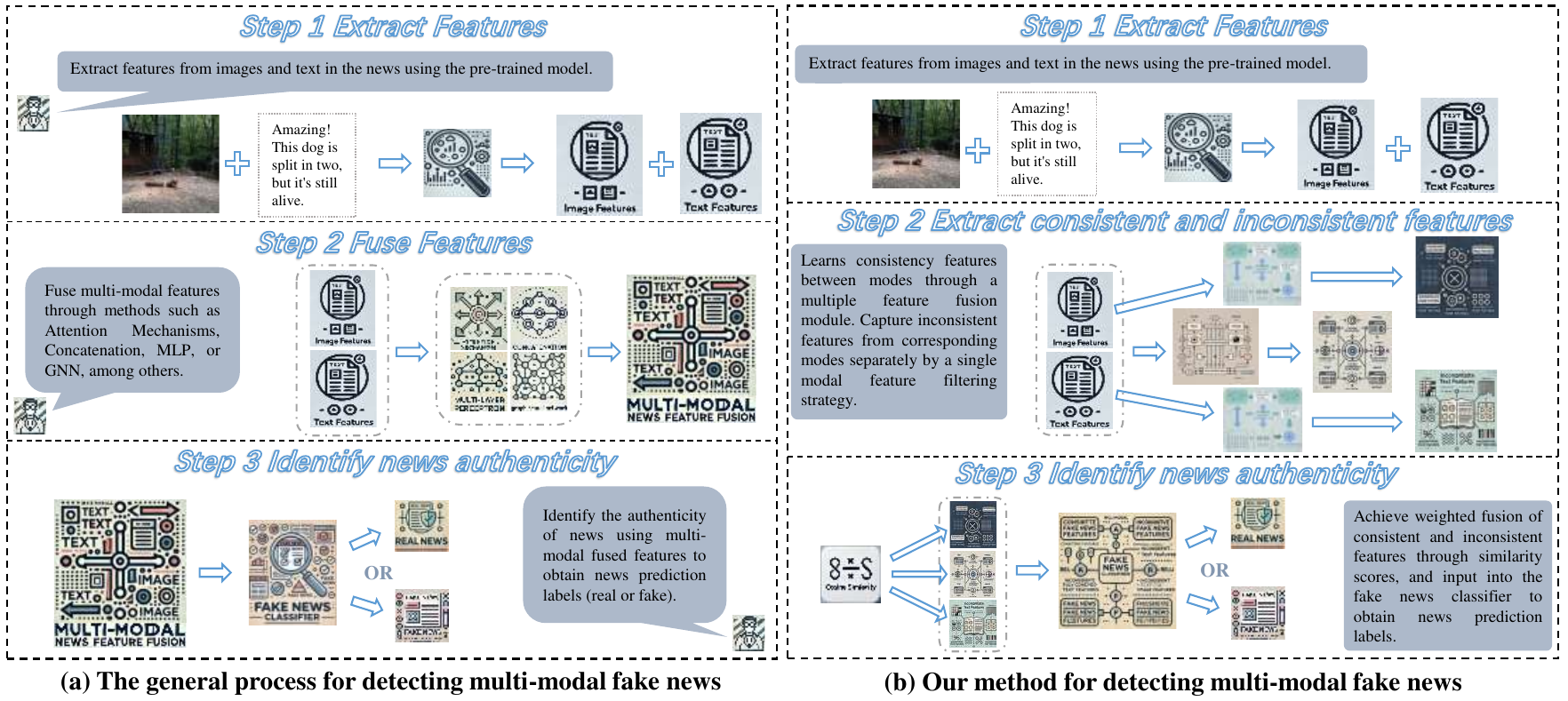}
\caption{Differences between existing methods and ours in fake news detection. Existing methods extract features from each modality and fuse these features to distinguish the authenticity of news. In contrast, our method simultaneously considers capturing inter-modal consistency features and inconsistency features within each modality, and adaptively adjusts the weights of each feature to enhance modal information interaction for detecting fake news.}
\label{detection-process}
\end{figure}

Fake news detection primarily focused on single modal content in news, such as text \cite{ma2016detecting,cheng2020vroc} or images \cite{jin2016novel,qi2019exploiting}. Early researchers found significant differences in the semantic styles between real and fake news. That is, fake news usually contain fewer transitive verbs and have a large number of emotional words, shorter words, and longer sentences. This study of single modal content used the semantic features of news and considered additional information such as writing style and image processing traces. Although single modal methods is effective in fake news detection, news content spread on social media platforms nowadays often includes multiple interconnected modalities, such as text and visuals. Such methods completely ignored the contribution of the remaining modal content and the correlation between various modalities. Subsequently, researchers shifted their attention to multi-modal feature fusion, which involved combining features from various modalities through concatenation, function mapping, etc. \cite{singhal2020spotfake+,zhou2020similarity}. These methods have some performance improvement compared to methods that only exploited single modal content. However, the multi-modal feature fusion approaches are overly simplistic, and the features of each modality are still independent, with no way to facilitate information exchange to promote inter-modal information understanding.

The general fake news detection process is shown in Fig. \ref{detection-process} (a), most fake news detection technologies used different models to extract high-level features from different modalities and integrated multi-modal features using attention mechanisms, concatenation or other methods. The fused multi-modal features are then fed into downstream classifiers for news prediction \cite{chen2022cross,zhou2023multi,wu2023human}. These technologies analyzed inter-modal consistency, i.e., aligning the semantics of images and text to generate fused features, thereby improving the accuracy of fake news detection. Meanwhile, Ying et al. \cite{ying2023bootstrapping} found that inter-modal consistency information does not necessarily play a critical role and even sometimes lead to incorrect decisions, especially easy to recognize news with high degree of text-image correlation as real news. However, many fake news can mislead readers precisely because of the high correlation between images and text. Moreover, since text and images in fake news may be fabricated, parts of the news that are less or not related between multiple modalities are more likely to contain clues inconsistent with common sense, which can be used to reveal the authenticity of the news. Therefore, while paying attention to the inter-modal consistency, the inter-modal inconsistency should also be considered, as these are two features of equal importance in fake news detection.

To address the aforementioned issues, we propose an innovative adaptive multi-modal feature fusion network that leverages the consistency and inconsistency information of multi-modal data to achieve fake news detection, as illustrated in Fig. \ref{detection-process} (b). When manually assessing the authenticity of a news, if the consistency of content across various modalities is high, we will pay more attention to whether there are inconsistencies or less relevant parts between modalities. Conversely, when the content across modalities is less relevant, we rely on the consistency between modalities to determine the news authenticity. Thus, MFF-Net extracts consistency information between modalities and inconsistency information within each modality, and promotes adaptive fusion of feature information through similarity weighting, thereby enhancing the detection performance of MFF-Net. Specifically, we first exploit the pre-trained SWIN Transformer (SWIN-T) and BERT to capture semantic information from images and texts. Meantime, the pre-trained CLIP is adopted to extract global information from images and texts. Then, we learn inter-modal consistency information through a multiple feature fusion module, and capture inconsistency information of corresponding modalities in two single modal branches. Finally, cosine similarity score of global features is calculated to adaptively adjust the use of each obtained detection information. The contributions are summarized as follows:

\begin{itemize}
    \item We design a multiple feature fusion module based on a common attention mechanism, which performs fine-grained fusion of semantic information from images and text through multiple pairs of common attention blocks. Meanwhile, it enhances the fused features using the semantic information from each modality to capture consistency features among multiple modalities in news. These consistency features is able to enhance information interaction between modalities, and promote information understanding.
    \item We devise a feature filtering strategy by employing inconsistency score vectors. Distinct from the existing approaches, where features of multiple modes are aggregated first and then the inconsistent features are extracted, we separately extract the inconsistent modal features of different mode by computing and reversing the similarity of semantic information of a single mode. By filtering out the useless noise and redundant information in the corresponding modal features, the utilization rate of the inconsistent features of a single mode can be enhanced and the phenomenon of mode information being masked can be effectively prevented.
    \item The proposed MFF-Net achieves better fake news detection performance in comparison with several state-of-the-art methods. Ablation experiments indicate that inconsistency information learned from each modality and consistency information captured between modalities both contribute positively to our MFF-Net.
\end{itemize}

\section{Network Architecture}
\begin{figure*}[t]
\centering
\includegraphics[width=0.64\textwidth]{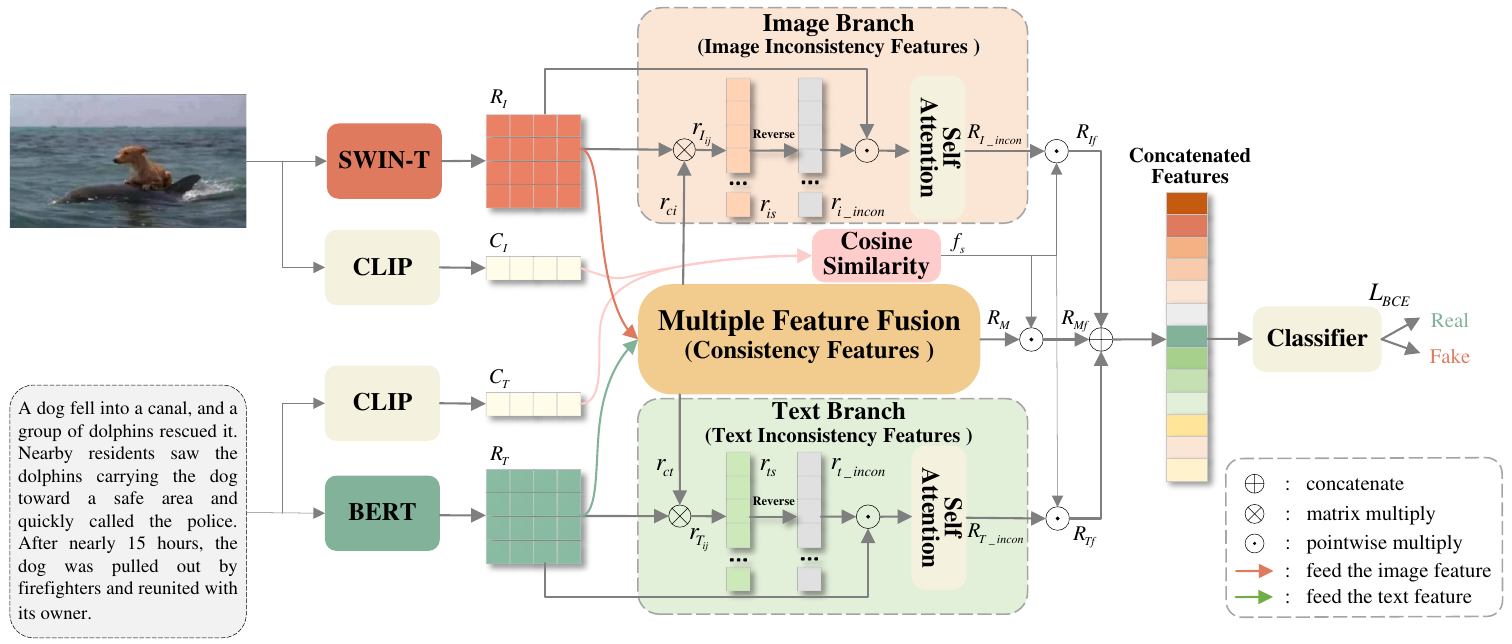} 
\caption{The overview of our MFF-Net. The SWIN-T, BERT, and CLIP models are employed to extract single modal features. Then, three parallel branches are utilized to extract the inconsistent and consistent information. Finally, the cosine similarity is computed to regulate the contribution degree of each feature, and the real and fake news are predicted by the classifier.}
\label{framework}
\end{figure*}

Our proposed MFF-Net enhances the accuracy of fake news detection through the adaptive fusion of the consistency and inconsistency features of multiple news modalities. The network architecture of MFF-Net is depicted in Fig. \ref{framework}. Firstly, three single modal feature extractors are employed to extract the semantic features $\{R_I, R_T\}$ and global features $\{C_I, C_T\}$ of images and texts. Subsequently, the image inconsistent feature $R_{I\_incon}$, text inconsistent feature $R_{T\_incon}$, and inter-modal consistency feature $R_M$ are learned via the image branch, text branch, and multiple feature fusion module, respectively. Then, the feature information is weighted by cosine similarity and fed into a classifier for fake news detection.

\subsection{Single modal Feature Extraction}
\textbf{Text modal semantic feature extraction:} We utilize the BERT pre-trained model \cite{devlin2018bert} to extract text semantic feature $R_T$ of news. BERT is a pre-trained natural language processing model based on the Transformer architecture. Its bidirectional approach allows it to consider both preceding and following text, enabling a more comprehensive and accurate understanding of the input. Additionally, BERT is pre-trained on a large corpus, acquiring rich language representations with strong transferability. The output from BERT's last hidden layer provides a detailed representation of context features, encompassing not only the semantic information of individual tokens but also the surrounding context. This makes BERT highly effective in understanding and processing text. Therefore, we use the BERT pre-trained model to process news text, and the output from BERT's last hidden layer is exploited as the extracted text semantic feature $R_T$.
Through this method, we can effectively extract deep semantic feature of news text, providing reliable text feature support for subsequent fake news detection.

\textbf{Image modal semantic feature extraction:} We use the SWIN-T pre-trained model \cite{liu2021swin} to extract image semantic feature $R_I$ of news. SWIN-T is a Transformer-based model for visual processing that introduces a hierarchical Transformer to extract features layer-by-layer, and captures both local and global features through a sliding window mechanism, which achieves superior performance in computer vision tasks such as image classification, object detection, and semantic segmentation. We introduce SWIN-T to handle the image modal of news, exploiting the output of the last Transformer layer as the image semantic feature $R_I$.
Through this method, we can effectively extract multi-scale features of news image, providing reliable visual feature support for fake news detection.

\textbf{Single modal global feature extraction:} We use the CLIP pre-trained model \cite{radford2021learning} to extract global features $C_I$ and $C_T$ for both image and text modalities simultaneously. The CLIP model achieves text and image embedding through multi-modal feature encoding, mapping them into a unified mathematical space. This feature provides significant advantages when calculating inter-modal consistency. CLIP has been pre-trained on a vast and diverse dataset and demonstrates strong robustness, especially in handling distribution variations and zero-sample learning scenarios. In fake news detection, CLIP can effectively be applied to detect unseen news data \cite{li2022language}. We choose to use the CLIP model to extract text global feature $C_T$ and image global feature $C_I$ for subsequent calculation of inter-modal consistency scores based on cosine similarity.
The similarity scores calculated from the two global features mapped to a unified mathematical space adaptively adjust the weights of each feature in fake news detection, which has a significant improvement effect on guiding the subsequent classifier to predict news authenticity.

\begin{figure}[t]
\centering
\includegraphics[width=0.4\textwidth]{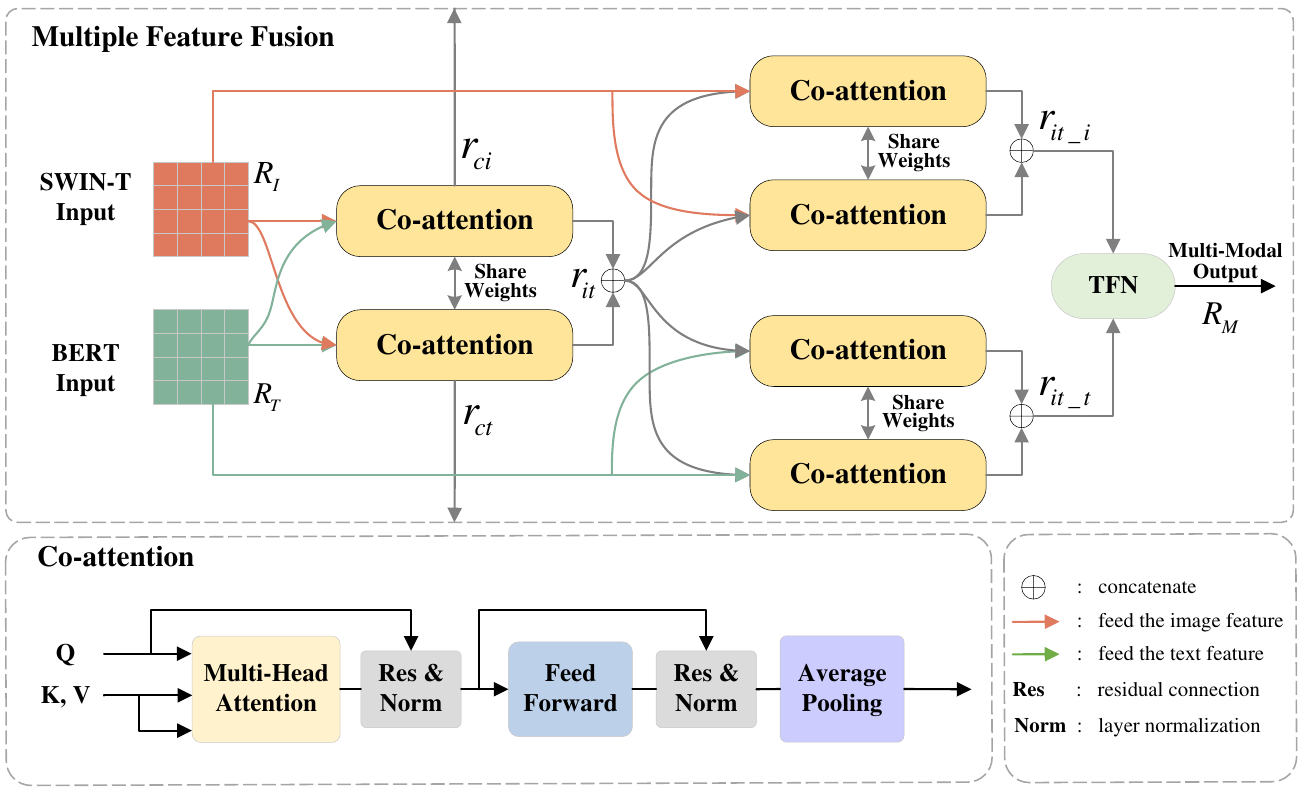} 
\caption{Illustration of our multiple feature fusion module. From left to right and from top to bottom, the first pair of Co-attention fuses the semantic features of text and image, the second and third pair of Co-attention are used to enhance the fused feature $r_{it}$, and the final inter-modal consistency information $R_M$ is obtained through the TFN fusion strategy.}
\label{fusion}
\end{figure}

\subsection{Interactive Learning of Inter-modal Consistency Information}
Since BERT and SWIN-T are not multi-modal models, there exists a considerable gap between text semantic feature $R_T$ and visual semantic feature $R_I$. If information fusion is conducted directly, the information between the modes cannot be interactively comprehended, and effective information between the modes cannot be learned. To address this issue, we proposed a multiple feature fusion module based on the Co-attention mechanism \cite{lu2019vilbert} for extracting inter-modal consistent features. Fig. \ref{fusion} depicts the structure of the multiple feature fusion module. The Co-attention mechanism mainly consists of a multi-head attention and feed forward layer, with residual connections and layer normalization behind each component. Through multi-head attention, it becomes feasible to focus on the valid information of text and image features simultaneously and establish associations between them. The feed forward layer can further process these correlation features and achieve more efficient information interaction when fusing different modal information. Residual connections and layer normalization can enhance the stability of training and strengthen the representation of features. As a result, by employing the Co-attention mechanism, the text feature $R_T$ extracted by BERT and the image feature $R_I$ extracted by SWIN-T can be effectively fused to enhance the understanding of inter-modal information and better learn inter-modal consistency features. Simultaneously, to prevent overfitting, we share weights between a pair of Co-attention blocks.

When fusing features through the first pair of Co-attention, the multi-head attention of each Co-attention block has $H$ heads. Mapping different modal features $R_T$ and $R_I$ to the same feature space by linear transformations, for each head $h$,
\begin{equation}\label{co-att-t}
\small
\left\{
\begin{aligned}
  Q^{h}_T&=R_TW^{T,h}_Q\quad
  K^{h}_T=R_TW^{T,h}_K\quad
  V^{h}_T=R_TW^{T,h}_V\\
  Q^{h}_I&=R_IW^{I,h}_Q\quad
  K^{h}_I=R_IW^{I,h}_K\quad
  V^{h}_I=R_IW^{I,h}_V
\end{aligned}
\right.
\end{equation}
where $Q^{h}_T,K^{h}_T,V^{h}_T \in \mathbb{R}^{n \times d_k}$, $Q^{h}_I,K^{h}_I,V^{h}_I \in \mathbb{R}^{p \times d_k}$, $d_k$ represents the dimensions of the query(Q), key(K), and value(V) vectors for each head in the multi-head attention. $W^{T,h}_Q,W^{T,h}_K,W^{T,h}_V \in \mathbb{R}^{d_t \times d_k}$ and $W^{I,h}_Q,W^{I,h}_K,W^{I,h}_V \in \mathbb{R}^{d_i \times d_k}$ are the learnable weight matrix for each head.

Then, the attention weight of a single attention head is calculated as follow:
\begin{equation}\label{att-weight-1}
\left\{
\begin{aligned}
  A^{h}_{TI}&=softmax(\frac{Q^{h}_T(K^{h}_I)^{T}}{\sqrt{d_k}})\\
  A^{h}_{IT}&=softmax(\frac{Q^{h}_I(K^{h}_T)^{I}}{\sqrt{d_k}})
\end{aligned}
\right.
\end{equation}
where $A^{h}_{TI} \in \mathbb{R}^{n \times p}$ and $A^{h}_{IT} \in \mathbb{R}^{p \times n}$.

The attention output of a single attentional head is calculated by the single attentional weight, which can be expressed as:
\begin{equation}\label{att-output-1}
\left\{
\begin{aligned}
  R^{h}_T&=A^{h}_{TI}V^{h}_I\\
  R^{h}_I&=A^{h}_{IT}V^{h}_T
\end{aligned}
\right.
\end{equation}

The output of $H$ attention heads in multi-head attention is concatenated and linear transformation is performed to obtain the final output feature of the multi-head attention.
\begin{equation}\label{final-out-1}
\left\{
\begin{aligned}
  R'_{T}&=concat(R^{1}_{T},R^{2}_{T}, \ldots , R^{H}_{T})W^{T}_{O}\\
  R'_{I}&=concat(R^{1}_{I},R^{2}_{I}, \ldots , R^{H}_{I})W^{I}_{O}
\end{aligned}
\right.
\end{equation}
where $W^{T}_{O} \in \mathbb{R}^{Hd_k \times d_t}$ and $W^{I}_{O} \in \mathbb{R}^{Hd_k \times d_i}$ are linear transformation matrices.

The features $R'_{T}$ and $R'_{I}$ are updated through residual connections and layer normalization. Finally, the updated feature is fed to the feed forward layer to strengthen the correlation between features, followed by residual connections, layer normalization, and average pooling. The strengthened features $r_{ct}$ and $r_{ci}$ are extracted as the output of each Co-attention block.
The output of the two Co-attention blocks is fused to obtain the first fused multi-modal feature $r_{it}$.

Some studies have demonstrated that information enhancement in multi-modal fake news detection can improve detection accuracy and robustness \cite{xiong2023trimoon}. To further assist MFF-Net in enhancing its comprehension of the information among the modalities of text and image and to increase the feature richness, we respectively enhance the information of the first integrated multi-modal feature $r_{it}$ with the text feature $R_T$ and the image feature $R_I$ through two pairs of Co-attention blocks ($CA$).
\begin{equation}\label{two-fuse-1}
\left\{
\begin{aligned}
  r_{it\_t}&=concat(CA(R_T,r_{it}),CA(r_{it},R_T))\\
  r_{it\_i}&=concat(CA(R_I,r_{it}),CA(r_{it},R_I))
\end{aligned}
\right.
\end{equation}

Meanwhile, the TFN fusion strategy \cite{zadeh2017tensor} can fully utilize various features for fusion. Hence, we employ the TFN strategy to integrate the two enhanced multi-modal features $r_{it\_t}$ and $r_{it\_i}$. Specifically, the dimension expansion of $r_{it\_t}$ and $r_{it\_i}$ is conducted with 1 respectively, and then the Cartesian product of $r_{it\_t}$ and $r_{it\_i}$ is calculated as follow,
\begin{equation}\label{product}
  R_M=[r_{it\_t};1] \bigotimes [r_{it\_i};1]
\end{equation}
Where ``$;$" is concatenation, and $\bigotimes$ is Cartesian product.

\subsection{Feature Filtering for Modal Inconsistency Information Learning}
Currently, the majority of fake news detection approaches based on multi-modal information directly concatenated and fused multiple single-modal information, causing information redundancy to a certain extent. Simultaneously, existing methods pay significantly less attention to the inconsistent information among modes compared to the consistent information between them. For the text branch and image branch depicted in Fig. \ref{framework}, we propose a single modal feature filtering strategy to extract inconsistent features from the image and text modes. This strategy initially computes the inconsistency score and filters the features of the text and image modes based on the self-attention mechanism \cite{vaswani2017attention} to eliminate redundant information and useless noise. It should be noted that redundant information refers to the highly consistent features of text and image extracted by the multiple feature fusion module, which is prone to overfitting. While useless noise refers to the information that has no positive impact on fake news detection within the two modal information. This strategy improves the utilization rate of inconsistent information and avoids the shading effect of the inconsistent information in one mode on that in another mode.

Taking image branch as an example, as shown in Fig. \ref{framework}, the similarity matrix $r_{I_{ij}}$ is obtained by multiplying the image semantic feature $R_I$ and the strengthened image feature $r_{ci}$ weighted with text attention.
In order to represent the inter-modal consistency score $r_{is}$ of each block in the image feature, the similarity matrix $r_{I_{ij}}$ is summed by column and then normalized by softmax.
The consistency score vector $r_{is}$ is then reversed to generate the inconsistency score vector, calculated as follows:
\begin{equation}\label{incon-score}
  r_{I\_incon}=1-r_{is}
\end{equation}
The higher the inconsistency score of the corresponding position, the less attention is paid to this area of the image during multiple feature fusion, which means that the corresponding position has inconsistency.

\begin{table*}[tb]
\begin{center}
\renewcommand{\tabcolsep}{2.6mm}
\renewcommand{\arraystretch}{0.6}
\begin{tabular}{@{}ccccccccc}
\noalign{\hrule height 1pt}
$\vphantom{W^{W^W}_{W_W}}$ \multirow{2}{*}{Dataset}&\multirow{2}{*}{Method}&\multirow{2}{*}{Accuracy}&\multicolumn{3}{c}{Fake news}&\multicolumn{3}{c}{Real news}\\
\cmidrule(r){4-6} \cmidrule(r){7-9}
&&&Precision&Recall&F1&Precision&Recall&F1\\
\noalign{\hrule height 0.5 pt}
\multirowcell{6}{Weibo}
&Spotfake+ (AAAI'20)&0.870&0.887&0.849&0.868&0.855&0.892&0.873\\
&CAFE (WWW'22)&0.840&0.855&0.830&0.842&0.825&0.851&0.837\\
&BMR (AAAI'23)&0.918&0.882&\textbf{0.948}&0.914&\textbf{0.942}&0.870&0.904\\
&MMFN (ICMR'23)&0.923&0.921&0.926&0.924&0.924&0.920&0.922\\
&CFFN (MM'23)&0.901&0.913&0.889&0.889&0.888&0.913&0.900\\
&MFF-Net (our)&\textbf{0.930}&\textbf{0.951}&0.909&\textbf{0.930}&0.910&\textbf{0.952}&\textbf{0.931}\\
\hline
\multirowcell{6}{Weibo-21}
&Spotfake+ (AAAI'20)&0.851&\textbf{0.953}&0.733&0.828&0.786&\textbf{0.964}&0.866\\
&CAFE (WWW'22)&0.882&0.857&0.915&0.885&0.907&0.844&0.876\\
&BMR (AAAI'23)&0.929&0.908&0.947&0.927&0.946&0.906&0.925\\
&MMFN (ICMR'23)&0.930&0.941&0.920&0.930&0.919&0.940&0.929\\
&CFFN (MM'23)&0.914&0.926&0.899&0.912&0.902&0.928&0.915\\
&MFF-Net (our)&\textbf{0.943}&0.934&\textbf{0.951}&\textbf{0.942}&\textbf{0.952}&0.935&\textbf{0.943}\\
\hline
\multirowcell{6}{GossipCop}
&Spotfake+ (AAAI'20)&0.858&0.732&0.372&0.494&0.866&0.962&0.914\\
&CAFE (WWW'22)&0.867&0.732&0.490&0.587&0.887&0.957&0.921\\
&BMR (AAAI'23)&0.895&0.752&0.639&0.691&0.920&0.965&0.936\\
&MMFN (ICMR'23)&0.894&0.799&\textbf{0.698}&0.684&0.910&0.964&0.936\\
&CFFN (MM'23)&0.885&0.749&0.610&0.672&0.894&0.959&0.925\\
&MFF-Net (our)&\textbf{0.911}&\textbf{0.838}&0.666&\textbf{0.742}&\textbf{0.924}&\textbf{0.969}&\textbf{0.946}\\
\noalign{\hrule height 1 pt}
\end{tabular}
\caption{Comparisons with SOTA methods on three real social media transmission news datasets. The highest value is \textbf{bold}.}
\label{SOTA}
\end{center}
\end{table*}

Then, the inconsistency score vector is weighted to the original image feature $R_I$, and the inconsistency feature representation is calculated. Since there may be some redundant information and interference noise in the obtained inconsistent feature, the self-attention mechanism (SA) is adopted to filter the obtained image inconsistent feature.
\begin{equation}\label{filter}
  R_{I\_incon}=SA(r_{I\_incon} \bigodot R_I)
\end{equation}
where $\bigodot$ is pointwise multiply.

As illustrated in the text branch, the text inconsistent feature $R_{T\_incon}$ can be captured in the same way as image inconsistent feature extraction.

\subsection{Weight Adaptation for News Authenticity Classification}
To better utilize the image inconsistency feature $R_{I\_incon}$, text inconsistency feature $R_{T\_incon}$, and inter-modal consistency feature $R_M$ in the classifier, we adjust the contribution weights of each feature in fake news detection by computing the consistency score of the text global feature $C_T$ and the image global feature $C_I$, guiding the classifier to learn useful information. Thus, the classifier can judge the truth of news more effectively.
Specifically, we calculate the cosine similarity $f_s$ of the image global feature $C_I$ and the text global feature $C_T$ extracted by the CLIP pre-trained model.
\begin{equation}\label{cos}
  f_s=\frac{C_T \cdot (C_I)^{T}}{\|C_T\| \cdot \|C_I\|}
\end{equation}

The key idea of MFF-Net is to imitate human identification of the truth and falseness of news, that is, when the degree of similarity between images and texts is high, the detection model pays more attention to the parts where they are inconsistent, and when the degree of similarity is low, the model focuses more on the parts where they are consistent. Therefore, we first normalize the cosine similarity score $f_s$ to be within the range of 0 and 1.
\begin{equation}\label{normfs}
  sim=\frac{1+f_s}{2}
\end{equation}

Then, we use $sim$ to weight the image inconsistency feature $R_{I\_incon}$, the text inconsistency feature $R_{T\_incon}$, and the inter-modal consistency feature $R_M$.
\begin{equation}\label{weightsim}
\left\{
\begin{aligned}
    R_{If}&=sim \cdot R_{I\_incon}\\
    R_{Tf}&=sim \cdot R_{T\_incon}\\
    R_{Mf}&=(1-sim) \cdot R_M
\end{aligned}
\right.
\end{equation}

Finally, the weighted inconsistency features and consistency feature are concatenated and fed into the fake news classifier, which consists of a fully connected layer and a ReLu activation function. The classifier will output the real-fake prediction result of the news.

The loss function of MFF-Net is binary cross entropy loss,
\begin{equation}\label{loss}
  L_{BCE}=-[ylog(y')+(1-y)log(1-y')]
\end{equation}
where $y$ and $y'$ denote the real and prediction label.

\begin{table*}[tb]
\begin{center}
\renewcommand{\tabcolsep}{2.6mm}
\renewcommand{\arraystretch}{0.6}
\begin{tabular}{@{}ccccccccc}
\noalign{\hrule height 1pt}
$\vphantom{W^{W^W}_{W_W}}$ \multirow{2}{*}{Dataset}&\multirow{2}{*}{Method}&\multirow{2}{*}{Accuracy}&\multicolumn{3}{c}{Fake news}&\multicolumn{3}{c}{Real news}\\
\cmidrule(r){4-6} \cmidrule(r){7-9}
&&&Precision&Recall&F1&Precision&Recall&F1\\
\noalign{\hrule height 0.5 pt}
\multirowcell{6}{Weibo}
&w/o ImageBranch&0.918&0.900&\textbf{0.944}&0.921&\textbf{0.939}&0.892&0.915\\
&w/o TextBranch&0.904&0.881&0.937&0.908&0.930&0.870&0.899\\
&w/o FeatureFusion&0.837&0.896&0.767&0.826&0.791&0.908&0.845\\
&w/o EnhanceFusion&0.909&\textbf{0.953}&0.862&0.905&0.871&\textbf{0.957}&0.912\\
&w/o Similarity&0.905&0.897&0.917&0.907&0.912&0.892&0.902\\
&MFF-Net&\textbf{0.930}&0.951&0.909&\textbf{0.930}&0.910&0.952&\textbf{0.931}\\
\hline
\multirowcell{6}{Weibo-21}
&w/o ImageBranch&0.927&0.917&0.935&0.926&0.936&0.918&0.927\\
&w/o TextBranch&0.912&0.871&\textbf{0.964}&0.915&\textbf{0.962}&0.862&0.909\\
&w/o FeatureFusion&0.844&0.796&0.920&0.854&0.909&0.772&0.835\\
&w/o EnhanceFusion&0.922&0.933&0.906&0.919&0.912&\textbf{0.938}&0.925\\
&w/o Similarity&0.919&0.925&0.909&0.917&0.913&0.929&0.921\\
&MFF-Net&\textbf{0.943}&\textbf{0.934}&0.951&\textbf{0.942}&0.952&0.935&\textbf{0.943}\\
\hline
\multirowcell{6}{GossipCop}
&w/o ImageBranch&0.901&0.800&0.646&0.715&0.919&0.961&0.940\\
&w/o TextBranch&0.882&0.683&\textbf{0.721}&0.701&\textbf{0.933}&0.920&0.926\\
&w/o FeatureFusion&0.823&0.544&0.495&0.518&0.882&0.901&0.891\\
&w/o EnhanceFusion&0.874&\textbf{0.859}&0.413&0.558&0.875&\textbf{0.984}&0.926\\
&w/o Similarity&0.887&0.746&0.626&0.681&0.914&0.949&0.931\\
&MFF-Net&\textbf{0.911}&0.838&0.666&\textbf{0.742}&0.924&0.969&\textbf{0.946}\\
\noalign{\hrule height 1 pt}
\end{tabular}
\caption{Comparisons of ablation study. ``w/o ImageBranch" denotes the detection model without the image branch, ``w/o TextBranch" denotes the detection model without the text branch, ``w/o FeatureFusion" denotes the detection model without the multiple feature fusion module, ``w/o EnhanceFusion" denotes the detection model that does not use single modal features to enhance the multi-modal fused feature in the multiple feature fusion module, and ``w/o Similarity" denotes the detection model that does not use the cosine similarity calculation. The highest value is \textbf{bold}.}
\label{ablation}
\end{center}
\end{table*}

\section{Experimental Results}
\subsection{Experimental Setup}
\textbf{Datasets:} We validate the performance of our MFF-Net on three publicly available datasets sourced from social medias: Weibo \cite{jin2017multimodal}, Weibo-21 \cite{nan2021mdfend}, and GossipCop \cite{shu2020fakenewsnet}. The Weibo dataset is the most widely used Chinese dataset for fake news detection, containing data from verified fake and real news on Sina Weibo between 2012 and 2016. The training set includes 3,783 real news and 3,675 fake news, while the testing set contains 1,685 news. The Weibo-21 dataset is an updated version of the Weibo dataset published in 2021, containing more recent Weibo news. This dataset includes 4,640 real news and 4,487 fake news, which we divided into a training set and testing set in a 9:1 ratio. GossipCop is an English dataset. The training set includes 7,974 real news and 2,036 fake news, and the testing set includes 2,285 real news and 545 fake news.

\textbf{Implement details:} We use the Adam optimizer for training, with a batch size set to 16, epochs set to 100, a learning rate set to 1. All experiments are conducted on a single NVIDIA RTX 4090 GPU.

\textbf{SOTA methods:} We compare MFF-Net with several state-of-the-art methods. That is, SpotFake+ \cite{singhal2020spotfake+}, CAFE \cite{chen2022cross}, BMR \cite{ying2023bootstrapping}, MMFN \cite{zhou2023multi}, and CFFN \cite{li2023cross}.

\subsection{Comparisons with SOTA Methods}
We use accuracy, precision, recall, and F1 score as evaluation metrics for fake news detection. Table \ref{SOTA} provides the performance comparison results with state-of-the-art methods. It can be seen that our MFF-Net achieves the highest detection accuracy and F1 score on each dataset, with most other evaluation metrics also ranking first or second. On the Weibo, Weibo-21, and GossipCop datasets, the detection accuracy of MFF-Net is improved by $0.7\%$, $1.3\%$, and $1.6\%$ compared to the best existing method, indicating the effectiveness of MFF-Net in fake news detection.

The superiority of our MFF-Net can be attributed to several reasons. Firstly, we calculate similarity scores based on global features extracted from the CLIP model to adaptively weight the fine-grained semantic features extracted from the BERT and SWIN-T models, which aligns with the human's habit of roughly judging whether multiple news modalities are consistent. Secondly, we extract inconsistency information for each single modal feature by filtering redundant information and interference noise, effectively preventing inconsistencies in one modality from being overwritten and ensuring that consistency and inconsistency information from each modality effectively participate in fake news detection. Finally, in the multiple feature fusion module, after merging the semantic features of image and text modalities, we further enhance the fused feature using the semantic features of the corresponding modality, which promotes the interactive understanding between modality information, and captures more effective consistency information.

\subsection{Ablation Study}
We perform an ablation study to validate the effectiveness of each module in the proposed MFF-Net. Table \ref{ablation} gives the comparison results. Comparing the experimental results of ``w/o ImageBranch", ``w/o TextBranch", and MFF-Net, it can be observed that the performance of ``w/o ImageBranch" and ``w/o TextBranch" has significantly decreased, indicating that a single modality's inconsistency information is insufficient to effectively support fake news detection. Meanwhile, the performance of ``w/o TextBranch" is slightly lower than that of ``w/o ImageBranch", suggesting that the inconsistency information from the text modality has a greater impact on fake news detection. Comparing the results of ``w/o FeatureFusion", ``w/o EnhanceFusion", and MFF-Net, it can be seen that ``w/o FeatureFusion" has the largest decrease in performance, indicating that the inter-modal consistency information extracted by the multiple feature fusion module significantly contributes to performance improvement, and inter-modal consistency information is crucial for fake news identification. Comparing ``w/o EnhanceFusion" with MFF-Net, ``w/o EnhanceFusion" shows a certain degree of performance drop, suggesting that our design of enhancing multi-modal fused feature with single modal features is effective, and also indicates that the interaction understanding of different modality information is helpful for extracting consistency information and improving detection performance. Comparing the results of ``w/o Similarity" with MFF-Net shows that adjusting the weights of inter-modal consistency feature and inconsistency features using cosine similarity score can guide the classifier in learning evidence clue, thereby improving fake news identification.

\section{Conclusion}
In this paper, we propose an adaptive multi-modal feature fusion network for fake news detection. MFF-Net strengthens the information interaction between news image and text modalities through the proposed multiple feature fusion module, thereby fully learning the inter-modal consistency information. Meanwhile, a single modal feature filtering strategy is designed to obtain inconsistency features from different modalities, which can filter out redundant information and interference noise, and avoid the mutual masking effect of multi-modal information. Furthermore, based on the global features of images and text, a similarity score is computed. Consistency and inconsistency features are weighted with the similarity score to adaptively fuse these features, so as to better guide the classifier in determining the authenticity of news. Experimental results demonstrate that our MFF-Net achieves greater performance in fake news detection compared with several SOTA methods.

\bibliography{aaai25}

\end{document}